\documentclass[letterpaper, 10 pt, conference]{ieeeconf}
\IEEEoverridecommandlockouts                              
\usepackage{comment}
\usepackage{url}

\usepackage{multirow}
\usepackage{graphicx}
\usepackage{epsfig}
\usepackage{epic,eepic}
\usepackage{times}
\usepackage{mathptmx}
\usepackage{cite}
\usepackage{color}

\usepackage{times}

\definecolor{red}{rgb}{1,0,0}
\definecolor{green}{rgb}{0,1,0}
\definecolor{blue}{rgb}{0,0,1}
\definecolor{violet}{rgb}{1,0,1}
\definecolor{cyan}{cmyk}{1,0,0,0}
\definecolor{magenta}{cmyk}{0,1,0,0}
\definecolor{yellow}{cmyk}{0,0,1,0}

\definecolor{white}{rgb}{1,1,1}

\newcommand{\CO}[1]{}

\newcommand{\CommentOut}[1]{}

 \newcommand{\editage}[1]{}

\begin{document}

% \twocolumn

\newcommand{\FIG}[3]{
\begin{minipage}[b]{#1cm}
\begin{center}
\includegraphics[width=#1cm]{#2}\\
{\scriptsize #3}
\end{center}
\end{minipage}
}

\newcommand{\FIGU}[3]{
\begin{minipage}[b]{#1cm}
\begin{center}
\includegraphics[width=#1cm,angle=180]{#2}\\
{\scriptsize #3}
\end{center}
\end{minipage}
}

\newcommand{\FIGm}[3]{
\begin{minipage}[b]{#1cm}
\begin{center}
\includegraphics[width=#1cm]{#2}\\
{\scriptsize #3}
\end{center}
\end{minipage}
}

\newcommand{\FIGR}[3]{
\begin{minipage}[b]{#1cm}
\begin{center}
\includegraphics[angle=-90,width=#1cm]{#2}
\\
{\scriptsize #3}
\vspace*{1mm}
\end{center}
\end{minipage}
}

\newcommand{\FIGRpng}[5]{
\begin{minipage}[b]{#1cm}
\begin{center}
\includegraphics[bb=0 0 #4 #5, angle=-90,clip,width=#1cm]{#2}\vspace*{1mm}
\\
{\scriptsize #3}
\vspace*{1mm}
\end{center}
\end{minipage}
}

\newcommand{\FIGCpng}[5]{
\begin{minipage}[b]{#1cm}
\begin{center}
\includegraphics[bb=0 0 #4 #5, angle=90,clip,width=#1cm]{#2}\vspace*{1mm}
\\
{\scriptsize #3}
\vspace*{1mm}
\end{center}
\end{minipage}
}

\newcommand{\FIGpng}[5]{
\begin{minipage}[b]{#1cm}
\begin{center}
\includegraphics[bb=0 0 #4 #5, clip, width=#1cm]{#2}\vspace*{-1mm}\\
{\scriptsize #3}
\vspace*{1mm}
\end{center}
\end{minipage}
}

\newcommand{\FIGtpng}[5]{
\begin{minipage}[t]{#1cm}
\begin{center}
\includegraphics[bb=0 0 #4 #5, clip,width=#1cm]{#2}\vspace*{1mm}
\\
{\scriptsize #3}
\vspace*{1mm}
\end{center}
\end{minipage}
}

\newcommand{\FIGRt}[3]{
\begin{minipage}[t]{#1cm}
\begin{center}
\includegraphics[angle=-90,clip,width=#1cm]{#2}\vspace*{1mm}
\\
{\scriptsize #3}
\vspace*{1mm}
\end{center}
\end{minipage}
}

\newcommand{\FIGRm}[3]{
\begin{minipage}[b]{#1cm}
\begin{center}
\includegraphics[angle=-90,clip,width=#1cm]{#2}\vspace*{0mm}
\\
{\scriptsize #3}
\vspace*{1mm}
\end{center}
\end{minipage}
}

\newcommand{\FIGC}[5]{
\begin{minipage}[b]{#1cm}
\begin{center}
\includegraphics[width=#2cm,height=#3cm]{#4}~$\Longrightarrow$\vspace*{0mm}
\\
{\scriptsize #5}
\vspace*{8mm}
\end{center}
\end{minipage}
}

\newcommand{\FIGf}[3]{
\begin{minipage}[b]{#1cm}
\begin{center}
\fbox{\includegraphics[width=#1cm]{#2}}\vspace*{0.5mm}\\
{\scriptsize #3}
\end{center}
\end{minipage}
}

% \bfseries
% \sffamily

\newcommand{\tabA}{
\begin{table}[t]
\caption{Performance results (achievement rate [\%]).}\label{tab:goal_achievement_rates}
\centering
\begin{tabular}{|l|r|r|r|}
\hline
\textbf{Workspace ID} & \textbf{Random (\%)} & \textbf{NNQL (\%)} & \textbf{MLP (\%)} \\
\hline
00800-TEEsavR23oF & 10.20 & 44.21 & 42.03 \\
\hline
00801-HaxA7YrQdEC & 8.10 & 29.20 & 27.12 \\
\hline
00802-wcojb4TFT3 & 12.60 & 50.82 & 54.11 \\
\hline
\end{tabular}
\end{table}
}

\newcommand{\figA}{
\begin{figure}[t]
\centering
\hspace*{1cm}\FIG{7}{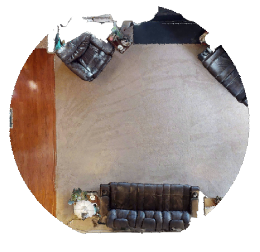}{}
\caption{%
We adopted a BEV (Bird's Eye View) omnidirectional local map, as shown in the figure, with the priority of being independent of specific platforms.
}\label{fig:A}
\end{figure}
}

\newcommand{\figB}{
\begin{figure}[t]
\centering
%\hspace*{1cm}%
\FIG{9}{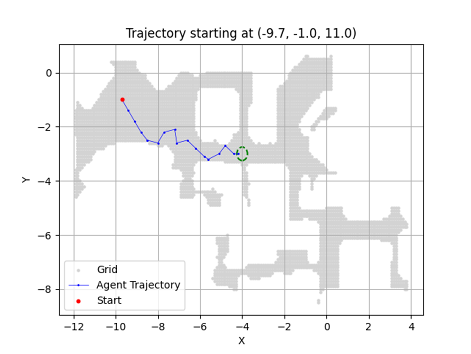}{}
\caption{%
Example of a robot's movement trajectory. The red dot represents the starting point, the blue curve represents the movement path, and the green circle represents the goal area.
}\label{fig:B}
\end{figure}
}

\title{\LARGE \bf
LMD-PGN: \\
Cross-Modal Knowledge Distillation from First-Person-View Images to Third-Person-View BEV Maps for Universal Point Goal Navigation
}

\author{Riku Uemura ~~~ Kanji Tanaka ~~~ Kenta Tsukahara ~~~ Daiki Iwata \thanks{Our work has been supported in part by JSPS KAKENHI Grant-in-Aid for Scientific Research (C) 20K12008 and 23K11270.}\thanks{$*$%
R. Uemura, K. Tanaka, K. Tsukahara, D. Iwata are with Robotics Coarse, 
Department of Engineering, University of Fukui, Japan. 
{\tt\small{\{ha210217@g., tnkknj@, mf240254@g., mf240050@g.\}u-fukui.ac.jp}}}}

\maketitle

\begin{abstract}
Point goal navigation (PGN) is a mapless navigation approach that trains robots to visually navigate to goal points without relying on pre-built maps. Despite significant progress in handling complex environments using deep reinforcement learning, current PGN methods are designed for single-robot systems, limiting their generalizability to multi-robot scenarios with diverse platforms. This paper addresses this limitation by proposing a knowledge transfer framework for PGN, allowing a teacher robot to transfer its learned navigation model to student robots, including those with unknown or black-box platforms.
We introduce a novel knowledge distillation (KD) framework that transfers first-person-view (FPV) representations (view images, turning/forward actions) to universally applicable third-person-view (TPV) representations (local maps, subgoals). The state is redefined as reconstructed local maps using SLAM, while actions are mapped to subgoals on a predefined grid. To enhance training efficiency, we propose a sampling-efficient KD approach that aligns training episodes via a noise-robust local map descriptor (LMD). Although validated on 2D wheeled robots, this method can be extended to 3D action spaces, such as drones.
Experiments conducted in Habitat-Sim demonstrate the feasibility of the proposed framework, requiring minimal implementation effort. This study highlights the potential for scalable and cross-platform PGN solutions, expanding the applicability of embodied AI systems in multi-robot scenarios.
\end{abstract}

\section{Introduction}

Point goal navigation (PGN), a rapidly emerging mapless navigation technology in the field of embodied AI, focuses on training robots to visually navigate safely and efficiently to a given goal point coordinate without relying on a-priori maps. With the support of recent advancements in datasets, simulators, and deep reinforcement learning technologies, PGN has remarkably improved to handle challenging setups, such as large-scale, multi-goal, dynamic, and unknown environments \cite{9687596}. However, current PGN technologies assume a single-robot system where the learner and user of the PGN model are identical, leaving it unclear whether it can generalize to multi-robot systems. In particular, their state-action maps are often specialized to the learner robot's own platform (sensors and actuators). While such models may generalize to a peer robot which is known at the time of learning, there is little chance that they can be generalized to unfamiliar robots with unknown platforms encountered after learning. 
To fully leverage the exceptional scalability and fault tolerance of multi-robot systems, it is imperative to establish a framework for transferring the PGN model trained by one robot (teacher) to others (students), including those on unknown and unfamiliar robot platforms.

This paper assumes that all robots can build, share, and interpret local maps \cite{bosse2003atlas}, and develops state-action maps that use local maps as an input modality, making them universally applicable to diverse robotic platforms, including black-box students unfamiliar at the time of PGN model training. Local maps, a fundamental component in robot navigation and SLAM, have been extensively studied in key applications such as mapping, self-localization, and path planning. Their integration into essential software like ROS further highlights their potential as a standard input modality for cross-robot platforms. Therefore, we argue that it is reasonable to assume that local maps are interpretable by robots (students) that were unfamiliar and unseen at the time of the teacher's PGN training. From the perspective of generality, the module for transferring the PGN model from the teacher to such a black-box student must be independent of both the teacher and the student, and it should be plug-and-play, ensuring easy integration. For example, in our experiment, the teacher and student are written in different programming languages (C++ and Python, respectively), and the teacher, student, and transfer module can be implemented on three different PCs, which ensures the requirements for independence and plug-and-play functionality.

We present a knowledge transfer framework formulated as a problem of knowledge distillation (KD) \cite{frosst2017distilling}, from a teacher PGN model with first-person-view images as input (FPV-PGN) to a student PGN model with third-person-view local map descriptors as input (TPV-PGN). Our contributions are summarized as follows:

(1) We propose a simple and effective approach to transfer FPV state-action representations (view images, turning/forward) specific to a robot's embodiment to universally applicable TPV representations (local maps, subgoals). Specifically, we convert 
a short-term FPV state-action sequence 
into a TPV representation with minimal information loss. 
In our solution, the state is redefined as a locally reconstructed map from short-term sequences using SfM/SLAM, and actions are redefined as points on a pre-defined regular grid on the local map.

(2) Second, we tackle the high training costs and scalability challenges of the PGN task by introducing a sampling-efficient KD scheme, framing teacher-student knowledge distillation as knowledge transfer from a reinforcement learning action planner (teacher) to an image action classifier (student). Furthermore, we propose a local map descriptor that aligns different episodes to a common reference coordinate system, along with techniques to enhance its noise invariance, enabling the summarization of multiple training episodes representing the same experience into a single episode. Although the experiments in this paper focus on a 2D wheeled visual robot, the proposed sampling efficiency is expected to be even more effective in 3D action spaces, such as drone formations.

(3) Finally, for proof-of-concept experiments, we consider a challenging scenario where the local map is represented as a 2D BEV image (BEV-PGN) and implement a practical prototype system. Our framework is easy to implement, requiring only a few dozen lines of code, suggesting high potential for future scalability. The effectiveness of the proposed method is evaluated through experiments using the publicly available photo-realistic simulator Habitat-Sim \cite{Habitat}.

\section{PGN Formulation}

This section formulates the PGN problem while addressing the computational complexity inherent in the task. In fact, PGN is computationally expensive, and improving its cost-effectiveness and scalability remains an ongoing area of research \cite{DBLP:conf/iclr/WijmansKMLEPSB20}.

\figB

\subsection{Robot Setup}\label{sec:embedding}

We consider a PGN task carried out by a wheeled robot equipped with a front-facing camera, specifically in the Habitat-Sim workspace (Figure \ref{fig:B}). PGN is a task in embodied AI \cite{gibson2014ecological} in which a robotic agent learns a state-action map to visually navigate to a specified goal point \cite{DBLP:conf/cvpr/KhannaRCYGCKCBM24}. This state-action map is most commonly formulated as a reinforcement learning model that takes a view image as the state input and outputs the next-best-action. In our framework, two types of input images, view images (\ref{sec:teacher}) and local maps (\ref{sec:student}), are considered. The output action consists of a turn of $\theta \in \Theta$ [deg] followed by a 0.5 m forward movement. $\Theta$ is the set of 13 turning angles $\{\theta_o, 0, \pm 30, \pm 60, \ldots, 180\}$, where $\theta_o$ represents a special turn action that randomly determines the turning angle, introduced to handle situations requiring fine adjustments in narrow passages.

\subsection{Embedding Model}\label{sec:embedding}

We assume the availability of an embedding model that maps input images to low-dimensional embedding vectors, 
which allows handling not only deep action planner models but also non-deep models,
as in \cite{DBLP:journals/ict-express/OhtaTY23}.   
Various methods exist for implementing such an embedding model (e.g., semantic vocabularies, appearance vocabularies), but investigating the optimal embedding is beyond the scope of this study; instead, we use a simple embedding model adopted in \cite{DBLP:journals/ict-express/OhtaTY23} for our experiments.
Specifically, in the offline training phase,  
(1) Using a prototype place classifier, the training workspace is divided into a $10 \times 5$ grid, and each grid cell is further divided with an angular resolution of 30 degrees. As a result, we obtain 600 place classes.  
Note that the GPS coordinates of each place class are not required.  
(2) Next, using class-specific image sets and categorical cross-entropy loss function, we train a VGG16 classifier, which takes an input image with 224$\times$224 format.  
Under this preparation, in the online embedding phase, the input image is mapped to a class-specific probability map using the place classifier, then transformed into a class-specific reciprocal rank embedding (RRE) vector, which is output as the final embedding vector.

\subsection{Annotations}

We assume the availability of an oracle planner as an annotator during the training phase only. Annotation here refers to the task where the oracle provides appropriate rewards (in the sense of reinforcement learning) for a robot's selected action at a given viewpoint. In our experiments, the shortest path from the viewpoint to the goal location on the aforementioned obstacle grid map is calculated using Dijkstra's algorithm, and the action toward the shortest path is defined as the best action. How these annotations are utilized in training is described in subsequent sections.

\subsection{Bumper Functionality}\label{sec:bumper}

We assume the availability of a bumper mechanism to prevent the robot from colliding with obstacles. This functionality is particularly important because, while the action planners (teacher/student) are designed to predict optimal actions toward the goal, even state-of-the-art methods are not flawless. Visual information alone often fails to ensure collision-free navigation in complex scenarios, such as narrow passages.

To mitigate this limitation, the bumper mechanism proactively eliminates dangerous action candidates that could lead to collisions. For instance, in our implementation, the trained action planner evaluates a reduced set of actions, scoring only those deemed safe by the bumper mechanism.

\figA

\subsection{Omni-Directional Local Map}

We assume, for simplicity, that the local maps used in the experiments are represented as BEV grid maps. Specifically, it is a BEV local map with a circular field of view (FOV) of a 2.5 m radius, as shown in Figure \ref{fig:A}. In practical applications, the local map may be reconstructed using methods such as simultaneous localization and mapping (SLAM) or structure from motion (SfM). Therefore, in such cases, the FOV is not only non-circular but could even be three-dimensional. The shape of the local map's FOV depends on the robot's local visual actions and the FOV of each viewpoint.
The issue of handling more realistic and diverse local maps is left as a future challenge, and this study does not address dependencies on specific robotic implementations, aiming to eliminate these dependencies as much as possible.

\section{Knowledge Transfer Method}

This section outlines the framework for teacher-to-student knowledge transfer, which includes the teacher model, the student model, and the process of teacher-student knowledge distillation.

\subsection{Teacher PGN Model}\label{sec:teacher}

We tackle the PGN problem by leveraging reinforcement learning (RL) \cite{sutton2018reinforcement}, a machine learning approach where an agent learns a state-action map through interactions with its environment. 
Specifically, this study utilizes Q-learning, a prominent RL technique. Q-learning seeks to estimate the expected cumulative reward $Q(s, a)$ for a robot executing action $a$ in state $s$. 
The state vector is an embedding vector, where the input local map is mapped to the embedding vector using the method described in Section \ref{sec:embedding}. 
Whenever the robot takes an action $a$ in state $s$, it receives an immediate reward $R$, and Q-learning updates the value of $Q(s, a)$ using the following iterative procedure:
\begin{equation}
Q(s,a) \leftarrow Q(s,a) + \alpha(R + \gamma \max_a Q(s', a)-Q(s,a)),
\end{equation}
where $\alpha = 0.1$ is the learning rate, $\gamma = 0.9$ is the discount factor, $R$ is the immediate reward (detailed below), and $s'$ is the next state. 

Oracle:
It is assumed that a BEV obstacle grid map recording the obstacle configuration in the workspace is available only during training. The robot's traversable region is represented as a grid map on the $x$-$y$ plane with a resolution of 0.1 m, allowing an oracle planner to clearly understand the permissible movement range within the workspace.

Reward:
The reward $R$ is computed as the difference between the shortest path costs at the episode's start and goal locations, determined using Dijkstra's algorithm with an oracle obstacle map as input.

Embedding:
All input images (view images or local maps) are preprocessed before being input to an RL model and dimensionally reduced into RRE vectors using the method described in \ref{sec:embedding}.

Training:
Nearest Neighbor Q-learning (NNQL) \cite{shah2018q}, which approximates $Q(s,a)$ using a nearest-neighbor search engine, is used as a function approximation method for $Q(s,a)$. Notably, naively implementing $Q(s,a)$ via a table indexed by state-action pairs is computationally infeasible for high-dimensional problems such as the one in this study.

\subsection{Student MLP Model}\label{sec:student}

The student model is designed to take a TPV local map as input and output the next-best-action. 

Architecture:
The student's architecture is implemented as a Multi-Layer Perceptron (MLP) \cite{white1963principles}. 
Its input is an embedding vector, where the input local map is mapped to the embedding vector using the method described in Section \ref{sec:embedding}. The MLP consists of two fully connected hidden layers with 2,048 and 1,024 neurons, respectively. Each hidden layer uses the ReLU (Rectified Linear Unit) activation function to introduce non-linearity and enable the model to learn complex patterns. The output layer consists of 13 neurons, corresponding to the 13 discrete actions available to the robot in the PGN task. A softmax activation function is applied to the output, generating a probability distribution over the possible actions.

Training:
Directly training the student MLP model is not the focus of this study, and the student model is trained only through teacher-student knowledge distillation, as described in Section \ref{sec:distill}.

\subsection{Teacher Student Knowledge Distillation}\label{sec:distill}

In existing knowledge distillation studies, teacher and student models often share the same architecture \cite{frosst2017distilling}. This similarity simplifies knowledge transfer and enhances the distillation effect. In contrast to these standard setups, this paper addresses the challenging setup of cross-architecture knowledge distillation, where the teacher and student models have different architectures. Further, the teacher model is a non-differentiable reinforcement learning action planner model (NNQL), while the student model is a differentiable image-based action classifier model (MLP). Few studies have explored cross-architecture knowledge distillation, especially cases like ours where the teacher and student models differ significantly.

Instead of assuming that class-specific probability maps are provided by the teacher as in typical knowledge distillation, 
we introduce a novel framework, conceptually similar to the one used in our previous work \cite{DBLP:conf/icra/TakedaT21}, where only class-specific rank values are provided.
The only difference between the standard knowledge distillation framework proposed by Hinton et al. and our framework is that we replace class-specific probability vectors with L1-normalized class-specific reciprocal rank vectors.
The Kullback-Leibler (KL) divergence loss is employed as the loss function to align the output probability distribution of the student model with that of the teacher model. The Adam optimizer with a learning rate of 0.001 is used for optimization. A batch size of 32 is adopted to balance computational efficiency and training stability.

\subsection{Invariant Reference Coordinate System}

The local map reconstructed from short-term visual experience is transformed into a rotation-invariant BEV reference coordinate using entropy minimization.
For entropy computation, the input image is first converted into the HSV color space, and the hue data is extracted. This hue data, represented as integer values in the range [0, M], where $M=255$. The histogram is L1-normalized to create an angle-specific histogram $p_j$, and the following entropy calculation formula is applied:
\begin{equation}
H = -\sum_{j=0}^M p_j \log_2 (p_j)
\end{equation}
where $p_j$ represents the probability of hue value $j$. This entropy calculation is applied to all pixels in the image, measuring the uniformity of features within the image.

As a result, the definition of place classes changes from combinations of position (50 types) and angles (12 types) (50 × 12 = 600 types) to positions (50 types) only. Consequently, the class-specific reciprocal rank vector becomes a 50-dimensional vector.
Notably, there is a clear many-to-one correspondence between the teacher model's 600 place classes and the student model's 50 place classes. 
During the knowledge distillation process, 
the 600-dimensional vector on the teacher side is compressed into a 50-dimensional vector. 
Specifically, the $i$-th element of the original vector $(i \in [0, 600))$ is mapped to 
the $j$-th element of the compressed vector $(j \in [0, 50))$ 
according to the formula $j = i / 12$. 
The KD loss is then calculated in the resulting 50-dimensional space.

\section{Experiments}

\subsection{Setup}

For the experiments, we utilized three workspaces 
00800-TEEsavR23oF, 00801-HaxA7YrQdEC, and 00802-wcojb4TFT35
from Habitat-Sim \cite{Habitat} by importing them into the simulator. 
The traversable area of each workspace is represented as a grid of at a 0.1 m resolution. 
The agent operates within this grid, starting from a randomly chosen initial position. The goal is to reach a specified target position. In each episode, the agent selects actions for a maximum of four steps, ending the episode upon reaching the goal. After the episode ends, the agent receives a reward and updates its Q-values based on the reward. The agent's actions are determined using an $\epsilon$-greedy method, balancing exploration and exploitation.

% Data Generator: A custom data generator, `NumpyArrayGenerator`, is used to supply data in batches, avoiding the need to load all data into memory at once. The batch size is set to 16, and the training data is shuffled.

$\epsilon$-greedy: 
The RL agent uses an $\epsilon$-greedy strategy to balance exploration and exploitation. During exploration, actions are chosen randomly, while during exploitation, the optimal action based on current Q-values is selected. The initial $\epsilon$ is set to 1.0 and gradually decreases to a minimum value of 0.1 as episodes progress. The decay rate varies based on the workspace's size, ensuring broad exploration in the early stages and optimal action selection in the later stages.

Training: The student MLP model is trained using KL divergence as the loss function, aiming to align the student model's output distribution with the target distribution generated by the teacher model. This distillation process efficiently transfers knowledge from the teacher model to the student model. Hyperparameters such as learning rate, epoch count, and mini-batch size are optimized to maximize the performance of the distillation method.

\tabA

\subsection{Results}

Table \ref{tab:goal_achievement_rates} presents the performance results using the goal achievement rate as the performance index. 
The goal achievement rate is defined as the ratio of test episodes in which the agent successfully reached an area 
within a radius of 0.25 meters from the goal, within 50 action steps.
For performance comparison, we use the following baselines: the ``random" planner, which decides actions randomly; the ``NNQL" planner, a teacher RL planner; and the proposed student ``MLP" planner.

From these results, the following conclusions can be drawn:
\begin{enumerate}
    \item NNQL demonstrated significantly better performance than the random method, confirming its effectiveness as a teacher model.
    \item The post-distillation MLP achieved goal achievement rates close to or exceeding that of NNQL, demonstrating the effectiveness of knowledge transfer through distillation.
    \item Specifically, in task 00802-wcojb4TFT3, the MLP outperformed NNQL, suggesting that dimensionality reduction and representation learning through distillation were effective.
    \item The significant differences in goal achievement rates across the three environments indicated that the robot's strengths and weaknesses vary depending on the workspace.
\end{enumerate}

This study marks the initial step toward inter-robot knowledge distillation for universal visual navigation, with plans to further develop this approach and extend its application to various domains in follow-up research.

\bibliography{reference} 

\begin{thebibliography}{10}

\bibitem{9687596}
Jiafei Duan, Samson Yu, Hui~Li Tan, Hongyuan Zhu, and Cheston Tan.
\newblock A survey of embodied ai: From simulators to research tasks.
\newblock {\em IEEE Transactions on Emerging Topics in Computational
  Intelligence}, 6(2):230--244, 2022.

\bibitem{bosse2003atlas}
Michael Bosse, Paul Newman, John Leonard, Martin Soika, Wendelin Feiten, and
  Seth Teller.
\newblock An atlas framework for scalable mapping.
\newblock In {\em 2003 IEEE International Conference on Robotics and Automation
  (Cat. No. 03CH37422)}, volume~2, pages 1899--1906. IEEE, 2003.

\bibitem{frosst2017distilling}
Nicholas Frosst and Geoffrey Hinton.
\newblock Distilling a neural network into a soft decision tree.
\newblock {\em arXiv preprint arXiv:1711.09784}, 2017.

\bibitem{Habitat}
Manolis Savva, Abhishek Kadian, Oleksandr Maksymets, Yili Zhao, Erik Wijmans,
  Bhavana Jain, Julian Straub, Jia Liu, Vladlen Koltun, Jitendra Malik, et~al.
\newblock Habitat: A platform for embodied ai research.
\newblock In {\em Proceedings of the IEEE/CVF international conference on
  computer vision}, pages 9339--9347, 2019.

\bibitem{DBLP:conf/iclr/WijmansKMLEPSB20}
Erik Wijmans, Abhishek Kadian, Ari Morcos, Stefan Lee, Irfan Essa, Devi Parikh,
  Manolis Savva, and Dhruv Batra.
\newblock {DD-PPO:} learning near-perfect pointgoal navigators from 2.5 billion
  frames.
\newblock In {\em 8th International Conference on Learning Representations,
  {ICLR} 2020, Addis Ababa, Ethiopia, April 26-30, 2020}. OpenReview.net, 2020.

\bibitem{gibson2014ecological}
James~J Gibson.
\newblock {\em The ecological approach to visual perception: classic edition}.
\newblock Psychology press, 2014.

\bibitem{DBLP:conf/cvpr/KhannaRCYGCKCBM24}
Mukul Khanna, Ram Ramrakhya, Gunjan Chhablani, Sriram Yenamandra,
  Th{\'{e}}ophile Gervet, Matthew Chang, Zsolt Kira, Devendra~Singh Chaplot,
  Dhruv Batra, and Roozbeh Mottaghi.
\newblock Goat-bench: {A} benchmark for multi-modal lifelong navigation.
\newblock In {\em {IEEE/CVF} Conference on Computer Vision and Pattern
  Recognition, {CVPR} 2024, Seattle, WA, USA, June 16-22, 2024}, pages
  16373--16383. {IEEE}, 2024.

\bibitem{DBLP:journals/ict-express/OhtaTY23}
Tomoya Ohta, Kanji Tanaka, and Ryogo Yamamoto.
\newblock Scene graph descriptors for visual place classification from noisy
  scene data.
\newblock {\em {ICT} Express}, 9(6):995--1000, 2023.

\bibitem{sutton2018reinforcement}
Richard~S Sutton.
\newblock Reinforcement learning: An introduction.
\newblock {\em A Bradford Book}, 2018.

\bibitem{shah2018q}
Devavrat Shah and Qiaomin Xie.
\newblock Q-learning with nearest neighbors.
\newblock {\em Advances in Neural Information Processing Systems}, 31, 2018.

\bibitem{white1963principles}
BW~White.
\newblock Principles of neurodynamics: Perceptrons and the theory of brain
  mechanisms, 1963.

\bibitem{DBLP:conf/icra/TakedaT21}
Koji Takeda and Kanji Tanaka.
\newblock Dark reciprocal-rank: Teacher-to-student knowledge transfer from
  self-localization model to graph-convolutional neural network.
\newblock In {\em {IEEE} International Conference on Robotics and Automation,
  {ICRA} 2021, Xi'an, China, May 30 - June 5, 2021}, pages 1846--1853. {IEEE},
  2021.

\end{thebibliography}
\bibliographystyle{unsrt}

\end{document}